# An Explainable Machine Learning Framework for Predicting Blood–Brain Barrier Permeability Using Molecular Descriptors

Fatemeh Mahmoudi

Department of Materials Science and Engineering, Sharif University of Technology, Azadi St., Tehran, Iran

**Abstract**

Blood–brain barrier (BBB) permeability is a critical determinant in the development of central nervous system therapeutics because it directly influences the ability of drug candidates to reach their target sites within the brain. In this study, an explainable machine learning framework was developed to predict BBB permeability using molecular descriptors generated from the MoleculeNet BBBP dataset with the RDKit cheminformatics toolkit. Fifteen physicochemical descriptors extracted from 2,039 compounds were used to train four supervised machine learning algorithms, including Logistic Regression, Support Vector Machine (SVM), Random Forest, and Extreme Gradient Boosting (XGBoost). Hyperparameter optimization was performed using GridSearchCV, while model interpretability was investigated using SHapley Additive exPlanations (SHAP). Among the evaluated models, the optimized XGBoost classifier achieved the best predictive performance, with an accuracy of 88.97%, a precision of 88.92%, a recall of 97.76%, an F1-score of 93.13%, and a ROC–AUC of 0.9282. Stratified five-fold cross-validation further demonstrated the robustness of the proposed model, yielding a mean ROC–AUC of 0.8982 ± 0.0130. Feature importance and SHAP analyses consistently identified TPSA, HBD, and LogP as the most influential molecular descriptors governing BBB permeability prediction. Overall, the proposed framework provides an accurate, interpretable, and computationally efficient approach for BBB permeability prediction and may serve as a valuable tool for the early-stage screening of CNS drug candidates.

## 1. Introduction

The blood–brain barrier (BBB) is a highly selective physiological barrier that regulates the transport of molecules between the bloodstream and the central nervous system (CNS) [1–3]. While the BBB protects the brain from toxins and pathogens, it also restricts the delivery of many therapeutic agents [2,4]. Consequently, accurate prediction of BBB permeability has become a critical step in the early stages of CNS drug discovery [5–7].

Experimental assessment of BBB permeability relies on specialized *in vitro* and *in vivo* assays, which are often expensive, labor-intensive, and time-consuming [8]. Consequently, computational prediction methods have emerged as efficient alternatives for prioritizing candidate molecules before experimental validation [9–11]. Quantitative structure–activity relationship (QSAR) models and machine learning (ML) techniques have demonstrated considerable potential for predicting BBB permeability using molecular descriptors derived from chemical structures [6,12,13].

In recent years, machine learning algorithms such as Logistic Regression [14], Support Vector Machine (SVM)[13], Random Forest (RF)[6,15], and Extreme Gradient Boosting (XGBoost)[16] have been successfully applied to various drug discovery tasks, including toxicity prediction, molecular property prediction, and BBB permeability classification. Although these models often achieve high predictive accuracy, many function as "black-box" approaches and provide limited insight into the molecular features driving their predictions[17]. This lack of interpretability may reduce confidence in their application to pharmaceutical research and rational drug design[12,18,19].

Explainable Artificial Intelligence (XAI) techniques address this limitation by revealing the contribution of individual molecular descriptors to model predictions[20]. Among these methods, SHapley Additive exPlanations (SHAP) has become one of the most widely adopted approaches for interpreting complex machine learning models[21,22] . Integrating explainability with predictive modeling enables researchers not only to achieve accurate predictions but also to gain mechanistic insight into the physicochemical properties governing BBB permeability [23,24].

Recent studies have demonstrated significant progress in the application of machine learning and explainable artificial intelligence for blood–brain barrier permeability prediction. Various computational approaches, including ensemble learning, graph neural networks, multimodal deep learning, and explainable machine learning frameworks, have been developed to improve predictive performance and enhance model interpretability. A summary of representative recent studies is presented in **Table *1***.

**Table 1. Summary of recent machine learning and explainable artificial intelligence approaches for blood–brain barrier permeability prediction**

| Reference | Dataset | Model | Explainability | Major finding |
|---|---|---|---|---|
| ACS Chem. Neurosci. (2024)[25] | BBB compounds | ML explainer models | Explainable AI | Molecular substructures associated with BBB penetration were identified. |
| Molecular Informatics (2024)[18] | 6665 compounds | SVM, XGBoost, ExtraTrees, DNN, GCN | SHAP | Molecular representation significantly affected prediction performance. |
| J. Phys. Chem. Lett. (2025)[23] | BBB dataset | Multimodal deep ensemble | SHAP, Attention | Multimodal learning improved both accuracy and interpretability. |
| Pharmaceutics (2026)[12] | 7807 compounds | Gradient boosting | SHAP, PDP | TPSA was identified as one of the most influential descriptors. |
| Biochemistry (2026)[24] | BBB molecules | Tree-based models, GNN | Explainable AI | Advanced molecular representations improved model interpretability. |

As shown in **Table *1***, recent studies have increasingly incorporated explainable artificial intelligence techniques, particularly SHAP, to improve the interpretability of BBB permeability prediction models. Ensemble learning methods, especially gradient boosting algorithms, have consistently demonstrated strong predictive performance across different datasets. In addition, recent investigations have highlighted the importance of advanced molecular representations, including molecular fingerprints, graph-based representations, and multimodal learning strategies. Despite these advances, developing computationally efficient and interpretable models remains an important challenge in BBB permeability prediction.

In this study, we developed an explainable machine learning framework for predicting BBB permeability using molecular descriptors generated by RDKit. Four supervised machine learning algorithms; Logistic Regression, Support Vector Machine, Random Forest, and XGBoost were systematically evaluated, followed by hyperparameter optimization of the XGBoost model. Model interpretability was investigated using SHAP analysis, while stratified five-fold cross-validation was employed to assess robustness and generalization ability. The proposed framework provides an accurate, interpretable, and computationally efficient approach for early-stage BBB permeability prediction and may serve as a valuable computational tool for supporting drug candidate screening.

## 2. Materials and Methods

### 2.1 Dataset

The Blood–Brain Barrier Penetration (BBBP) dataset from the MoleculeNet benchmark collection was used in this study. The dataset contains experimentally validated small molecules annotated according to their ability to penetrate the blood–brain barrier. Molecular structures were represented as Simplified Molecular Input Line Entry System (SMILES) strings together with corresponding binary labels indicating BBB permeability. After data preprocessing, a total of **2,039 compounds** were retained for analysis. This dataset served as the basis for molecular descriptor calculation, machine learning model development, and performance evaluation.

### 2.2 Molecular Descriptor Calculation

Molecular descriptors were calculated from the SMILES representations using the RDKit cheminformatics library. A total of fifteen physicochemical descriptors were selected to characterize the structural and physicochemical properties of each molecule. These descriptors included Molecular Weight (MW), Exact Molecular Weight, LogP, Topological Polar Surface Area (TPSA), Molecular Refractivity (MolMR), Number of Hydrogen Bond Donors (HBD), Number of Hydrogen Bond Acceptors (HBA), Number of Rotatable Bonds, Ring Count, Heavy Atom Count, Number of Heteroatoms, Fraction of $sp^3$ Carbon Atoms (FractionCSP3), Number of Valence Electrons, Number of Aromatic Rings, and Number of Aliphatic Rings. Invalid or non-processable molecular structures were excluded prior to descriptor calculation. The resulting descriptor matrix was subsequently used as the input feature set for machine learning model development.

### 2.3 Exploratory Data Analysis

Exploratory data analysis (EDA) was performed to evaluate the quality and distribution of the calculated molecular descriptors before model development. The dataset was examined for missing values, descriptor completeness, and data consistency. Summary statistics were generated for all descriptors, and no missing values were identified following descriptor calculation. The distribution of each descriptor was also visually inspected to identify potential outliers. In addition, the distribution of BBB-permeable and non-permeable compounds was examined to evaluate class balance within the dataset. These analyses confirmed that the dataset was suitable for machine learning model training without requiring additional preprocessing or data imputation.

### 2.4 Machine Learning Models

Four supervised machine learning algorithms were evaluated for BBB permeability prediction, including Logistic Regression (LR), Support Vector Machine (SVM), Random Forest (RF), and Extreme Gradient Boosting (XGBoost). Logistic Regression served as a baseline linear classifier, whereas SVM represented a kernel-based learning approach. Random Forest and XGBoost were selected because of their ability to capture complex nonlinear relationships among molecular descriptors and their demonstrated predictive performance in cheminformatics applications. The dataset was randomly divided into training and testing subsets using an **80:20 train–test split** with a fixed random seed to ensure reproducibility.

### 2.5 Model Evaluation

The predictive performance of each machine learning model was assessed using five commonly employed classification metrics, including Accuracy, Precision, Recall, F1-score, and the Area Under the Receiver Operating Characteristic Curve (ROC–AUC). Confusion matrices and ROC curves were further generated to evaluate classification performance from complementary perspectives. Among these metrics, ROC–AUC was considered the primary performance indicator because it provides a threshold-independent measure of classification capability.

### 2.6 Hyperparameter Optimization

To improve predictive performance, the XGBoost classifier was optimized using GridSearchCV. A stratified five-fold cross-validation strategy was employed during the optimization process to ensure reliable parameter selection. Different combinations of hyperparameters, including the number of estimators, maximum tree depth, learning rate, subsampling ratio, and column sampling ratio, were systematically explored. The parameter combination that achieved the highest mean cross-validation ROC–AUC score was selected as the final optimized model.

### 2.7 SHAP Explainability

Model interpretability was investigated using SHapley Additive exPlanations (SHAP). SHAP values were calculated to quantify the contribution of individual molecular descriptors to model predictions. Both global and local interpretation techniques were employed, including SHAP summary plots and SHAP waterfall plots. These explainability analyses provided valuable insights into the molecular descriptors that most strongly influenced BBB permeability prediction while improving the transparency of the optimized machine learning model.

### 2.8 Cross-Validation

The robustness and generalization capability of the optimized XGBoost model were further evaluated using stratified five-fold cross-validation. Stratification preserved the original class distribution within each validation fold, thereby providing a reliable estimate of model performance on unseen data. The mean ROC–AUC together with its standard deviation across all validation folds was calculated to assess prediction stability, model robustness, and resistance to overfitting.

# 3. Results

## 3.1 Molecular Descriptor Analysis

**Table 2. Top molecular descriptors ranked according to the Random Forest feature importance analysis**

| Rank | Molecular Descriptor | Importance |
|---|---|---|
| 1 | TPSA | 0.1569 |
| 2 | NumHeteroatoms | 0.1057 |
| 3 | LogP | 0.1003 |
| 4 | HBD | 0.0896 |
| 5 | ExactMolWt | 0.0735 |
| 6 | MolecularWeight | 0.0698 |
| 7 | MolMR | 0.0654 |
| 8 | HBA | 0.0597 |
| 9 | FractionCSP3 | 0.0558 |
| 10 | NumValenceElectrons | 0.0498 |
| 11 | RotatableBonds | 0.0459 |
| 12 | HeavyAtomCount | 0.0422 |
| 13 | NumAliphaticRings | 0.0382 |
| 14 | RingCount | 0.0251 |
| 15 | NumAromaticRings | 0.0221 |

Feature importance analysis was performed using the Random Forest model to identify the molecular descriptors contributing most to BBB permeability prediction. The feature importance scores are summarized in **Table *2***, while their graphical representation is shown in **Figure *1***.

TPSA was identified as the most influential molecular descriptor (importance = 0.1569), followed by NumHeteroatoms (0.1057), LogP (0.1003), Hydrogen Bond Donors (HBD, 0.0896), and Exact Molecular Weight (ExactMolWt, 0.0735). These descriptors are closely associated with molecular polarity, lipophilicity, molecular size, and hydrogen-bonding capacity, all of which are well-established determinants of passive diffusion across the blood–brain barrier. The observed feature ranking is therefore consistent with established physicochemical principles governing BBB transport and supports the reliability of the developed prediction model.

### 3.2 Model Performance

**Table 3. Performance comparison of the evaluated machine learning models for BBB permeability prediction**

| Model | Accuracy | Precision | Recall | F1-score | ROC-AUC |
|---|---|---|---|---|---|
| Logistic Regression | 0.8358 | 0.8490 | 0.9551 | 0.8989 | 0.8866 |
| Support Vector Machine | 0.8652 | 0.8620 | 0.9808 | 0.9175 | 0.9117 |
| Random Forest | 0.8848 | 0.8863 | 0.9744 | 0.9282 | 0.9163 |
| XGBoost | 0.8873 | 0.8935 | 0.9679 | 0.9292 | 0.9192 |
| Optimized XGBoost | 0.8897 | 0.8892 | 0.9776 | 0.9313 | 0.9282 |

Four supervised machine learning algorithms together with an optimized XGBoost model were evaluated for BBB permeability prediction. Their predictive performance is summarized in **Table *3***. Logistic Regression served as the baseline classifier and achieved an accuracy of 83.58% with a ROC–AUC of 0.8866. Support Vector Machine improved the overall performance and achieved the highest recall (98.08%), indicating excellent sensitivity in identifying BBB-permeable compounds. Random Forest further improved predictive accuracy and demonstrated robust classification performance. Among all evaluated models, the optimized XGBoost classifier achieved the best overall performance, yielding an accuracy of 88.97%, a precision of 88.92%, a recall of 97.76%, an F1-score of 93.13%, and a ROC–AUC of 0.9282. These results indicate that ensemble learning algorithms consistently outperformed the conventional machine learning models for BBB permeability prediction.

### 3.3 Feature importance Analysis

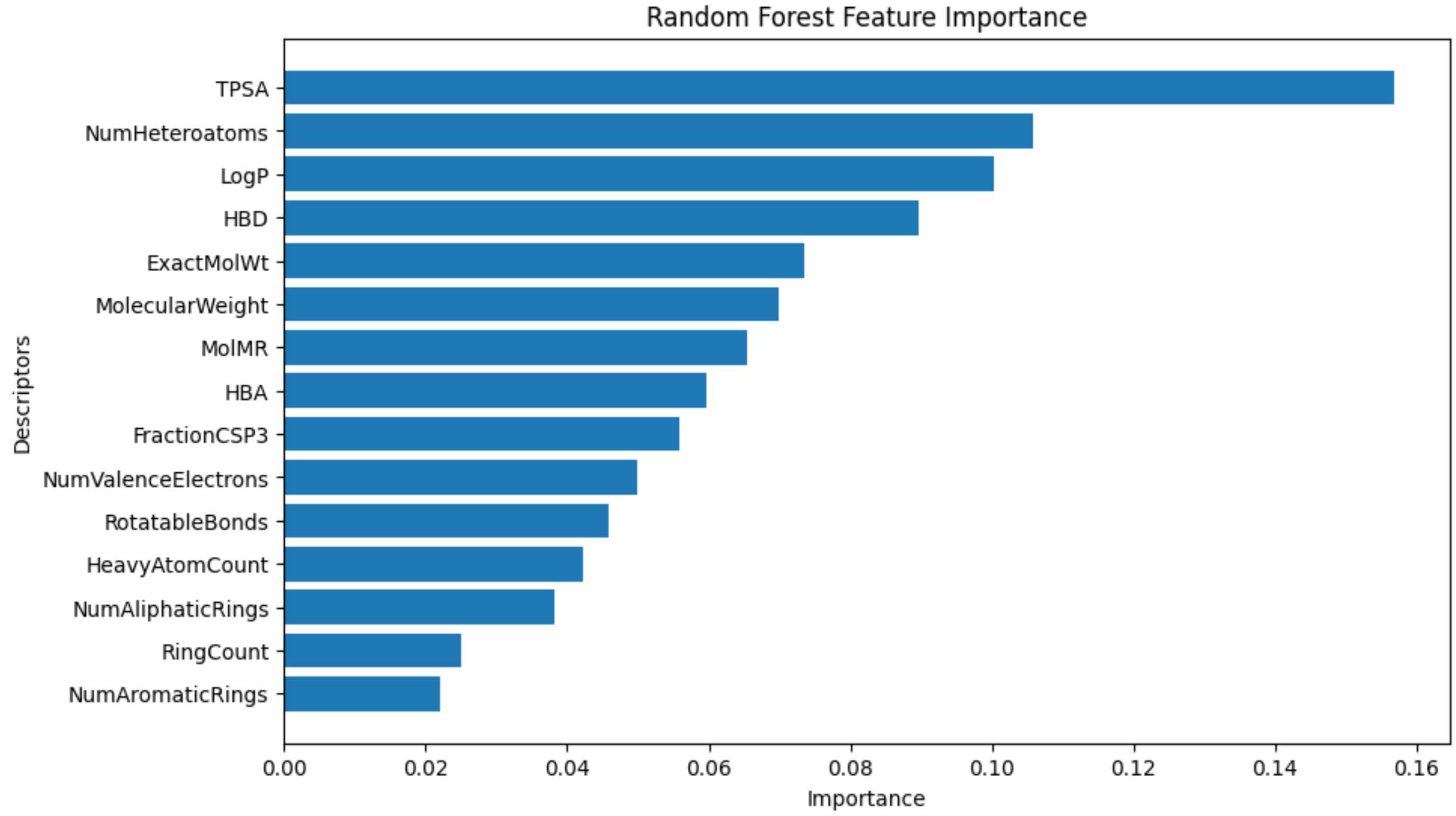

**Figure 1. Random Forest feature importance ranking of the molecular descriptors used for BBB permeability prediction. Topological Polar Surface Area (TPSA) was identified as the most influential descriptor, followed by the number of heteroatoms, LogP, hydrogen bond donors (HBD), and molecular weight–related descriptors. Higher feature importance values indicate a greater contribution of the corresponding descriptor to the predictive performance of the Random Forest model.**

Feature importance analysis was performed using the Random Forest model to identify the molecular descriptors contributing most to BBB permeability prediction (**Figure *1***). TPSA was identified as the most influential descriptor, followed by HBD, NumHeteroatoms, LogP, and molecular weight–related descriptors. These findings indicate that molecular polarity, hydrogen-bonding capacity, and lipophilicity play dominant roles in determining BBB permeability. In contrast, descriptors such as Ring Count and NumAromaticRings contributed comparatively less to the predictive performance of the model.

### 3.4 SHAP Interpretation

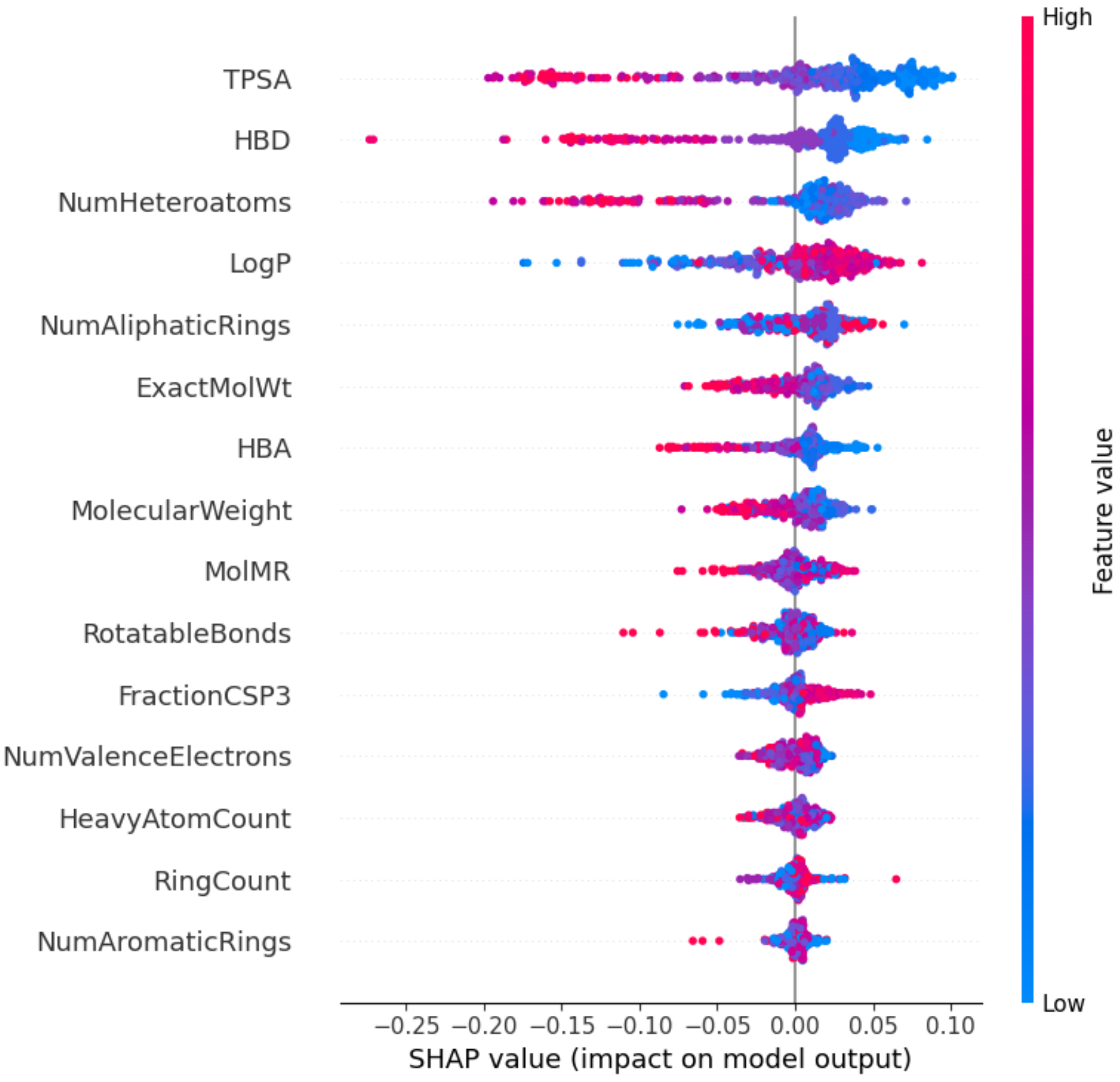

**Figure 2. SHAP summary plot illustrating the global contribution of molecular descriptors to the optimized XGBoost model. Each point represents an individual compound, and its horizontal position indicates the SHAP value, reflecting the contribution of that descriptor to the predicted BBB permeability. Red points correspond to high descriptor values, whereas blue points indicate low descriptor values. TPSA, HBD, NumHeteroatoms, and LogP were identified as the most influential descriptors. Lower TPSA and HBD values generally increased the predicted probability of BBB permeability, while higher LogP values were associated with increased BBB penetration.**

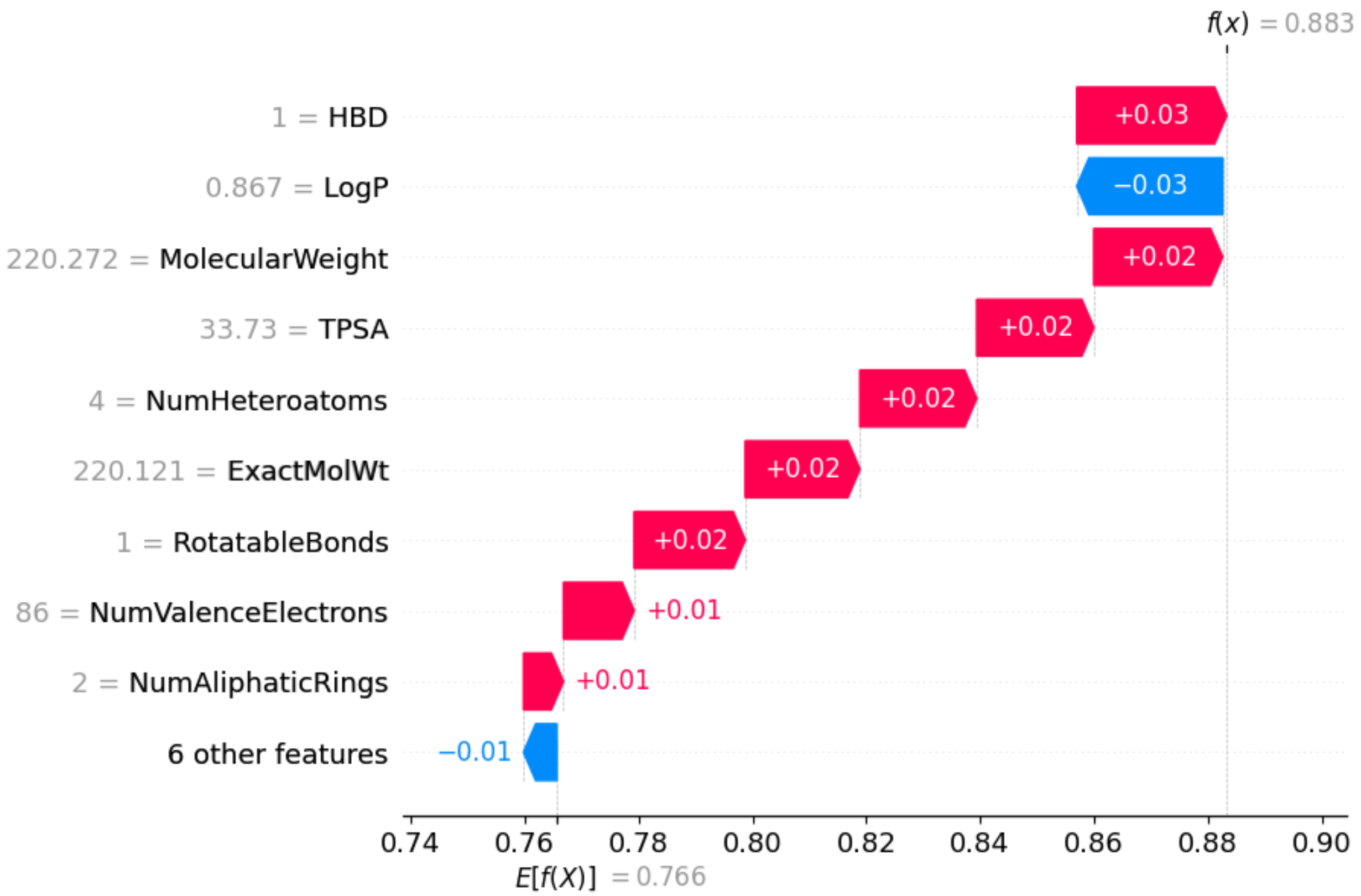

**Figure 3. SHAP waterfall plot explaining the prediction of a representative compound by the optimized XGBoost model. Starting from the model's baseline prediction (E[f(X)] = 0.766), individual molecular descriptors either increased (red) or decreased (blue) the predicted BBB permeability. The cumulative contributions of these descriptors resulted in a final model output of 0.883.**

SHAP analysis was performed to provide a global interpretation of the optimized XGBoost model (**Figure *2***). The SHAP summary plot ranked molecular descriptors according to their contribution to the model output. TPSA was identified as the most influential feature, followed by HBD, NumHeteroatoms, and LogP. The distribution of SHAP values indicated that compounds with lower TPSA and fewer hydrogen bond donors generally contributed positively to BBB permeability predictions, whereas higher lipophilicity (LogP) tended to increase the predicted probability of BBB penetration. These observations are consistent with the established physicochemical principles governing passive diffusion across the blood–brain barrier.

To provide a local interpretation of the optimized XGBoost model, a SHAP waterfall plot was generated for a representative compound (**Figure *3***). The prediction started from the model's baseline output (0.766) and was progressively adjusted according to the contribution of individual molecular descriptors. In this representative example, HBD, MolecularWeight, TPSA, NumHeteroatoms, ExactMolWt, and RotatableBonds positively influenced the predicted BBB permeability, whereas LogP slightly reduced the prediction. The cumulative effect of these descriptors increased the final prediction score to **0.883**, illustrating how individual molecular properties collectively influenced the model's decision for a single compound.

### 3.5. Receiver Operating Characteristic Analysis

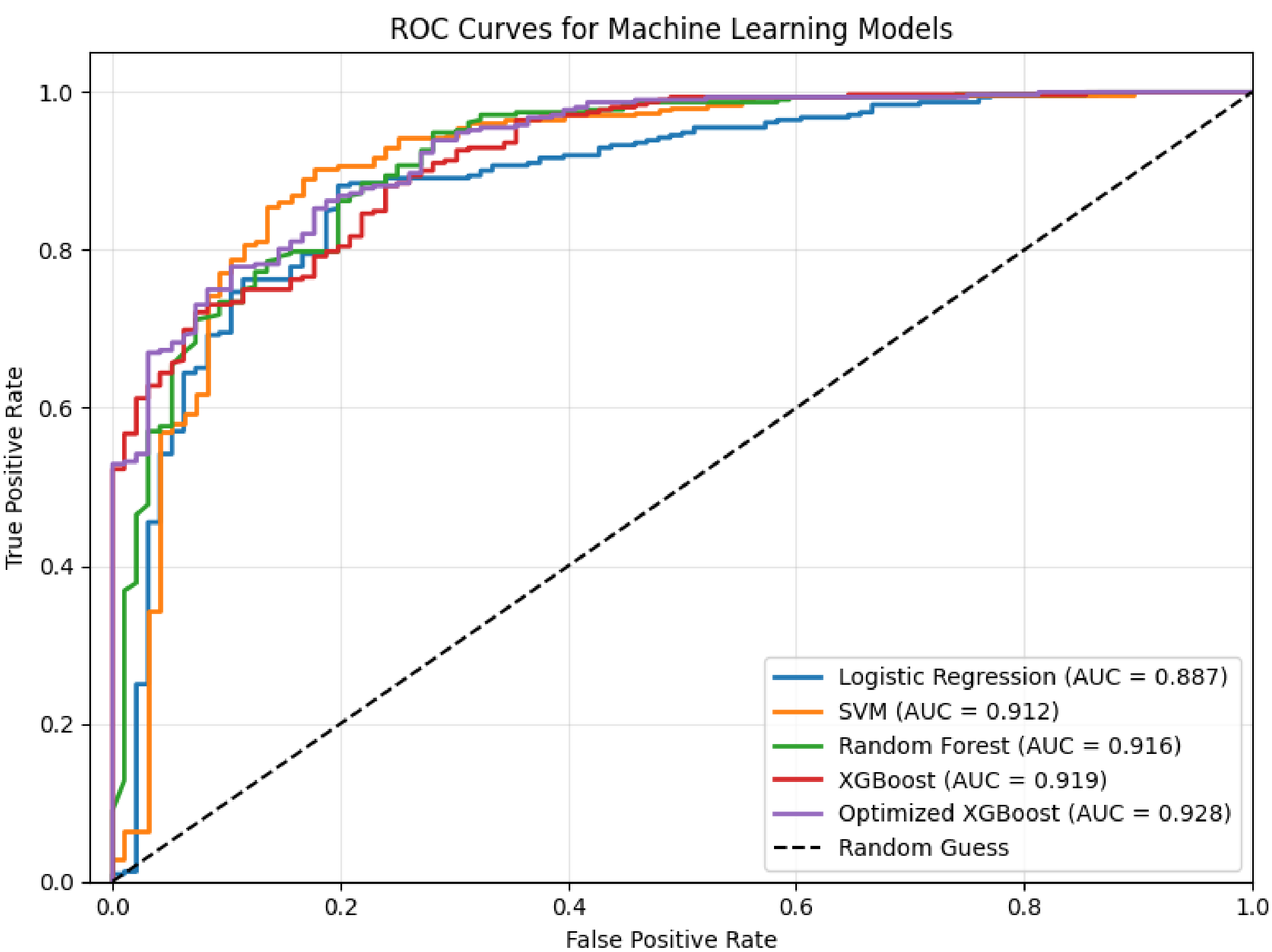

**Figure 4. Receiver Operating Characteristic (ROC) curves comparing the performance of Logistic Regression, Support Vector Machine (SVM), Random Forest, XGBoost, and the optimized XGBoost model for BBB permeability prediction. The optimized XGBoost model achieved the highest discriminative performance with a ROC–AUC of 0.9282, followed by XGBoost (0.9192), Random Forest (0.9163), SVM (0.9117), and Logistic Regression (0.8866). The dashed diagonal line represents the performance of a random classifier.**

The discriminative performance of the evaluated machine learning models was assessed using Receiver Operating Characteristic (ROC) curve analysis (**Figure** ***4***). The optimized XGBoost classifier achieved the highest ROC–AUC (**0.9282**), followed by XGBoost (**0.9192**), Random Forest (**0.9163**), Support Vector Machine (**0.9117**), and Logistic Regression (**0.8866**). Overall, the ensemble learning models consistently outperformed the conventional machine learning algorithms, demonstrating their superior ability to capture the nonlinear relationships between molecular descriptors and BBB permeability.

### 3.6. Confusion Matrix Analysis

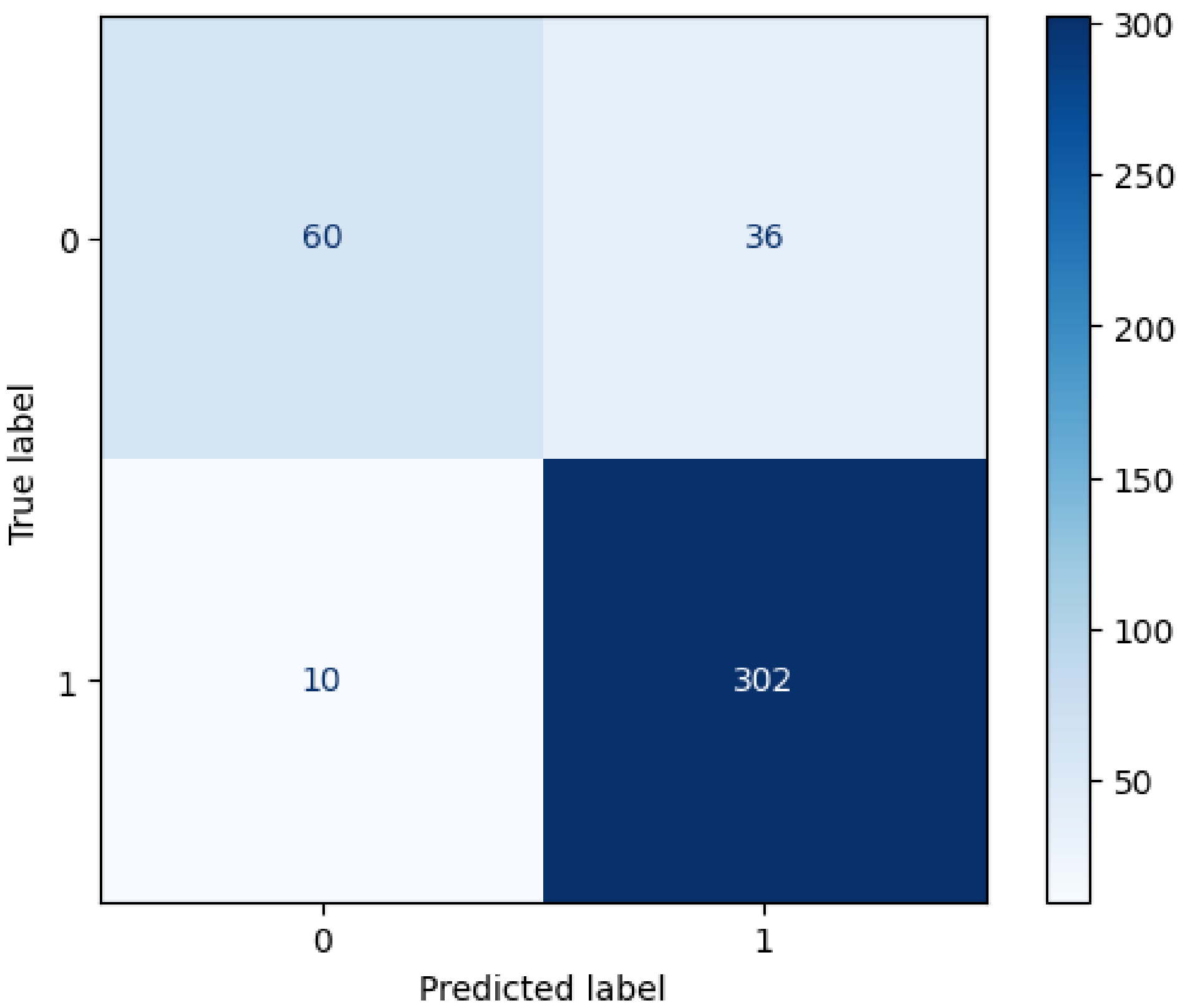

**Figure 5. Confusion matrix of the optimized XGBoost classifier for BBB permeability prediction. The model correctly classified 302 BBB-permeable compounds and 60 non-permeable compounds, while misclassifying 10 false negatives and 36 false positives, demonstrating strong overall classification performance.**

The classification performance of the optimized XGBoost model was further evaluated using the confusion matrix (**Figure 5**). The model correctly classified **302** BBB-permeable compounds (true positives) and **60** non-permeable compounds (true negatives). Only **10** BBB-permeable compounds were incorrectly classified as non-permeable (false negatives), whereas **36** non-permeable compounds were misclassified as permeable (false positives). These findings demonstrate that the optimized XGBoost classifier achieved high sensitivity while maintaining satisfactory specificity, supporting its effectiveness for BBB permeability prediction.

### 3.7 Cross-Validation Results

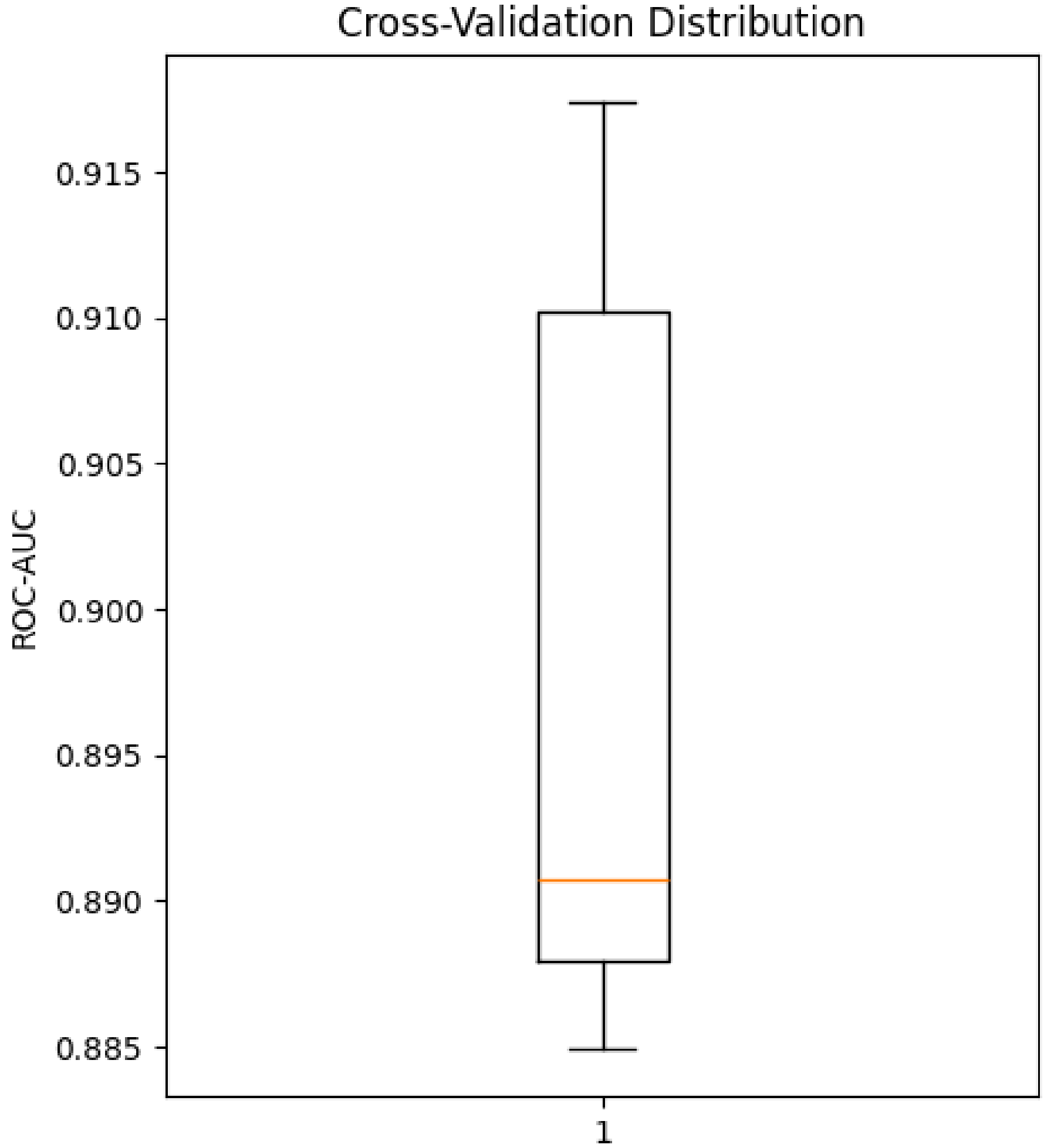

**Figure 6. Boxplot showing the distribution of ROC–AUC scores obtained from stratified five-fold cross-validation of the optimized XGBoost model. The narrow interquartile range and limited variability indicate stable and consistent predictive performance across different validation folds.**

The robustness of the optimized XGBoost model was further evaluated using stratified five-fold cross-validation. The distribution of ROC–AUC scores across the validation folds is presented in **Figure *6***, while the individual fold results are summarized in **Table *4***. The model achieved a mean ROC–AUC of 0.8982 ± 0.0130, with consistently high performance across all validation folds. The relatively small standard deviation, together with the narrow distribution of ROC–AUC values, indicates good model stability, robust generalization capability, and low sensitivity to variations in the training–testing data partitions.

### 3.8 Cross-Validation Results

**Table 4. Stratified five-fold cross-validation results of the optimized XGBoost model**

| Fold | ROC-AUC |
|---|---|
| Fold 1 | 0.8849 |
| Fold 2 | 0.8907 |
| Fold 3 | 0.9102 |
| Fold 4 | 0.8879 |
| Fold 5 | 0.9174 |
| Mean ± SD | 0.8982 ± 0.0130 |

The robustness and generalization capability of the optimized XGBoost model were evaluated using stratified five-fold cross-validation. As summarized in **Table 4**, the model achieved a mean ROC–AUC of **0.8982 ± 0.0130**, demonstrating stable predictive performance across different validation folds. The relatively small standard deviation indicates that the optimized model generalized well and maintained consistent classification performance across different data partitions.

### 3.9. Overall Model Performance

Overall, the ensemble learning algorithms outperformed the conventional machine learning methods for BBB permeability prediction. Logistic Regression provided acceptable baseline performance but exhibited the lowest predictive accuracy among the evaluated models. Support Vector Machine demonstrated excellent sensitivity by achieving the highest recall, while Random Forest and XGBoost produced superior overall classification performance. After hyperparameter optimization, the XGBoost classifier achieved the highest ROC-AUC (0.9282), indicating its superior capability to discriminate between BBB-permeable and non-permeable compounds. These findings suggest that gradient boosting algorithms are particularly effective for capturing the nonlinear relationships between molecular descriptors and BBB permeability.

## 4. Discussion

Prediction of blood–brain barrier (BBB) permeability remains a challenging task in drug discovery because BBB transport is governed by multiple physicochemical factors, including molecular size, polarity, lipophilicity, hydrogen-bonding capacity, and molecular topology. Previous studies have demonstrated that conventional experimental methods for evaluating BBB permeability are often labor-intensive and costly, which has increased the demand for computational approaches based on machine learning and quantitative structure–activity relationship (QSAR) models. Recent reviews have further emphasized the growing role of machine learning in accelerating the early-stage screening of CNS drug candidates.

In the present study, four supervised machine learning algorithms and an optimized XGBoost classifier were evaluated using fifteen molecular descriptors generated by RDKit. Among all investigated models, the optimized XGBoost classifier achieved the highest predictive performance, with a ROC–AUC of 0.9282 and an F1-score of 0.9313. These findings are consistent with previous investigations demonstrating that ensemble learning and boosting-based algorithms effectively capture the complex nonlinear relationships between molecular descriptors and BBB permeability. Liang et al.[18] reported that ensemble and graph-based models achieved excellent predictive performance for BBB permeability prediction, whereas Tiwari et al.[12] showed that gradient boosting algorithms outperformed several conventional machine learning approaches when applied to experimentally validated BBB datasets.

Feature importance analysis (**Figure *1***) identified Topological Polar Surface Area (TPSA) as the most influential molecular descriptor (importance = 0.1569), followed by NumHeteroatoms (0.1057), LogP (0.1003), Hydrogen Bond Donors (HBD, 0.0896), and Exact Molecular Weight (0.0735). These findings strongly agree with previous studies that identified molecular polarity and lipophilicity as the primary determinants of BBB permeability. In particular, Tiwari et al.[12] also reported TPSA as the most influential descriptor in their SHAP analysis, while Liang et al.[18] demonstrated that molecular representations describing polarity and lipophilicity substantially influenced model predictions. Lower TPSA values generally facilitate membrane permeation by reducing polar interactions with aqueous environments, whereas appropriate lipophilicity improves passive diffusion across biological membranes. Similarly, hydrogen-bonding capacity directly affects the ability of compounds to penetrate the BBB, explaining the substantial contribution of HBD observed in the present study.

To further improve model interpretability, SHAP analysis was performed. The SHAP summary plot (**Figure *2***) provided a global explanation of descriptor contributions across the entire dataset and confirmed the dominant influence of TPSA, HBD, LogP, and NumHeteroatoms. In contrast, the SHAP waterfall plot (**Figure *3***) provided a local explanation by illustrating how individual descriptors collectively affected the prediction of a representative compound. These findings are consistent with recent studies emphasizing the importance of explainable artificial intelligence for BBB prediction. Liang et al.[18] employed SHAP to identify influential molecular descriptors, whereas Yang et al.[24] demonstrated that explainable artificial intelligence substantially improves the transparency of BBB prediction models. Similarly, recent reviews have highlighted model interpretability as a critical requirement for the practical application of machine learning in pharmaceutical research.

The predictive capability of the optimized XGBoost model was further confirmed by the confusion matrix (**Figure *5***), which demonstrated a favorable balance between sensitivity and specificity. Notably, the relatively small number of false negatives reduced the likelihood of overlooking potentially BBB-permeable compounds during the early stages of drug discovery. Furthermore, stratified five-fold cross-validation (**Figure *6***) demonstrated stable predictive performance, with a mean ROC–AUC of $0.8982 \pm 0.0130$. The limited variability among validation folds suggests that the proposed model generalized well and was not strongly dependent on a specific data partition. Similar validation strategies have been widely adopted in recent BBB prediction studies to evaluate model robustness and reduce the risk of overfitting.

Despite the encouraging results, several limitations should be acknowledged. First, the present study relied exclusively on two-dimensional molecular descriptors generated by RDKit and did not incorporate molecular fingerprints, three-dimensional conformational descriptors, or graph-based molecular representations. Previous studies have demonstrated that integrating multiple molecular representations can further improve prediction performance. Second, external validation using an independent dataset was not performed. Tiwari et al.[12] demonstrated that external validation significantly improves model reliability and generalizability. Therefore, future studies may incorporate graph neural networks, molecular fingerprints, multimodal learning strategies, and independent validation datasets to further improve both prediction accuracy and model interpretability.

## 5. Conclusion

In this study, an explainable machine learning framework was developed for predicting blood–brain barrier (BBB) permeability using molecular descriptors generated by RDKit. Four supervised machine learning algorithms, including Logistic Regression, Support Vector Machine, Random Forest, and XGBoost, were systematically evaluated.

Among all investigated models, the optimized XGBoost classifier achieved the best predictive performance, with an accuracy of 88.97%, an F1-score of 93.13%, and a ROC–AUC of 0.9282. Feature importance analysis and SHAP interpretation consistently identified Topological Polar Surface Area (TPSA), Hydrogen Bond Donors (HBD), molecular lipophilicity (LogP), and the number of heteroatoms as the most influential descriptors governing BBB permeability.

Furthermore, stratified five-fold cross-validation confirmed the robustness and generalization ability of the proposed model. The integration of explainable artificial intelligence improved model transparency by providing both global and local interpretations of molecular descriptor contributions.

Although the present study relied exclusively on two-dimensional molecular descriptors and did not include external validation, the proposed framework demonstrated that combining machine learning with explainable artificial intelligence provides an accurate and interpretable approach for BBB permeability prediction. Future studies may incorporate molecular fingerprints, graph neural networks, multimodal molecular representations, and independent validation datasets to further improve prediction performance and model generalizability.

The proposed framework may serve as a practical computational tool for accelerating the early-stage screening and prioritization of CNS drug candidates while simultaneously improving the interpretability of BBB permeability prediction models.

**Data and Code Availability**

The datasets, source code, processed molecular descriptors, and supplementary materials supporting the findings of this study are publicly available at: https://github.com/Aydafzs/BBB-Permeability-Prediction-ML